\documentclass[11pt,a4paper]{article}
\usepackage[hyperref]{naaclhlt2019}
\usepackage{times}
\usepackage{latexsym}
\usepackage{graphicx}
\usepackage{url}

\aclfinalcopy 

\title{UM-IU@LING at SemEval-2019 Task 6: Identifying Offensive Tweets Using BERT and SVMs}

\author{Jian Zhu\\ 
  Department of Linguistics \\
  University of Michigan \\
  Ann Arbor, MI, USA \\
  {\tt lingjzhu@umich.edu} \\\And
  Zuoyu Tian\\ 
  Department of Linguistics \\
  Indiana University \\
  Bloomington, IN, USA \\
  {\tt zuoytian@iu.edu} \\ \And
  Sandra K\"{u}bler\\
  Department of Linguistics \\
  Indiana University \\
  Bloomington, IN, USA \\
  {\tt  skuebler@indiana.edu} \\}

\date{}

\begin{document}
\maketitle
\begin{abstract}
This paper describes the UM-IU@LING's system for the SemEval 2019 Task 6: Offens\-Eval. We take a mixed approach to identify and categorize hate speech in social media. In subtask A, we fine-tuned a BERT based classifier to detect abusive content in tweets, achieving a macro F$_1$ score of 0.8136 on the test data, thus reaching the 3rd rank out of 103 submissions. In subtasks B and C, we used a linear SVM with selected character $n$-gram features. For subtask C, our system could identify the target of abuse with a macro F$_1$ score of 0.5243, ranking it 27th out of 65 submissions.  
\end{abstract}

\section{Introduction}
\label{intro}
With the increased influence of social media on modern society, large amounts of user-generated content emerge on the internet. Besides the exchange of ideas, we also see an exponential increase of aggressive and potentially harmful content, for example, hate speech. If we consider the amount of user-generated data, it is impractical to manually identify the malicious speech. Thus we need to develop methods to detect offensive speech automatically through computational models. However, this task is challenging because natural language is fraught with ambiguities, and language in social media is extremely noisy. Here we present our method to automatically identifying offensive content in tweets. 

We primarily focus on detecting whether a tweet contains offensive content or not (subtask A), and then determining the target of the offensive content (subtask C). For subtask A, we use pre-trained word embeddings by fine-tuning the BERT model \cite{devlin2018bert} for detecting offensive tweets. For subtasks B and C, BERT did not perform well, either because of limited training data or because we did not find the appropriate hyperparameters. Thus we use an SVM classifier with character $n$-grams as features. We accidentally flipped the predicted labels in our submission to subtask B, which is why we do not report results of subtask B here.
Among all teams participating in OffensEval, our models ranks 3rd out of 103 on subtask A and 27th out of 65 on subtask C. \cite[see][]{offenseval}.

\section{Related Work}

Detecting offensive language online is becoming more and more important  \cite{schmidt2017survey,founta2018large,malmasi2018challenges}. To build an effective classifier, one of the major problems is to find the appropriate features. Normally, two types of features are utilized:  surface features like $n$-grams and word representations trained by neural network. Most offensive language classifiers are trained on different types of surface features with approaches like SVM~\cite{malmasi2018challenges,arroyo2018cyberbullying}, Random Forest~\cite{burnap2015cyber}, and Logistic Regression~\cite{davidson2017automated}. Recently, word embeddings trained in neural networks have been shown to achieve good performance in offensive language identification tasks~\cite{badjatiya2017deep}.  Benchmarks of the first shared task on aggression identification \cite{kumar2018benchmarking} show that half of the top 15 systems are trained on neural networks. 
 
Using pre-trained word embeddings for feature extraction has been shown to be highly effective in multiple NLP tasks. Traditional word embeddings are extracted from shallow neural networks trained on a large swathes of texts required to learn the contextual representations of words. Examples include skip-grams \citep{mikolov2013distributed} and GloVe \citep{pennington2014glove}. However, these embeddings are learned from an aggregation of all possible word contexts, which may gloss over semantic nuances in representations.

Recent models like ELMo \cite{peters2018deep} and BERT \cite{devlin2018bert} significantly advanced the state-of-the-art in language modeling by learning context-sensitive representations of words. ELMo goes beyond word embeddings by learning representations that are functions of the entire input sentence \cite{peters2018deep}. However, ELMo is still considered shallow with two bidirectional LSTM layers, and more recent transformer based language models such as the OpenAI Generative Pre-trained Transformer (GPT) \citep{radford2018improving} and Bidirectional Encoder Representations from Transformers (BERT) \citep{devlin2018bert} have been extended to a depth of up to twelve layers. The OpenAI GPT is still a unidirectional language model while BERT is trained to be bidirectional with two novel prediction tasks, Masked LM and Next Sentence Prediction. The pre-trained BERT model has been shown to give significant improvements in a series of downstream tasks over ELMo and OpenAI GPT \citep{devlin2018bert}.

 However, identifying offensive language is not a simple task. Challenges during identification include but are not limited to the fact that surface language features fail to capturing subtle semantic difference, and the shortage of undisputed annotated data \cite{malmasi2018challenges}. Most of previous studies focus on distinguishing between offensive and  non-offensive  language \cite{kwok2013locate,djuric2015hate}, 
which is the goal of subtask A in the current shared task. But part of challenge consists of the intertwined nature of such messages having negative connotations and profanity. 
\citet{dinakar2011modeling} show that it is important to tease these two factors apart. 
\citet{malmasi2018challenges} first address the issue of distinguishing hate speech from general profanity.

\section{Methodology and Data}
The subtasks in the shared task are rather different. In subtask A, the goal is to identify offensive tweets; in subtask B and C, the aim is to distinguish targeted and untargeted offense and to classify the targeted ones into different types. Subtask A requires sensitivity to subtle changes in word meaning in context while the other subtasks are more categorical in nature. However, both suffer from data sparsity. 
Therefore, we decided, backed by empirical validation on the trial data, to utilize different methods for the subtasks, namely, BERT embeddings for subtask A, and an SVM classifier for subtasks B and C. 

The data collection method used to compile the dataset in OffensEval is described by \newcite{OLID}. We  used the official training data and trial data provided by the shared task to train the classifier. Our implmentations can be found at: https://github.com/zytian9/SemEval-2019-Task-6.

\subsection{Subtask A: Identifying Abusive Content}
The goal of subtask A is to identify whether a tweet contains offensive content by training a model to perform binary classification. There are 13,240 tweet instances in the training data, in which each instance has been labeled as either 'offensive' ('OFF') or 'not offensive' ('NOT'). The model takes a tweet as input and predicts the corresponding label of that tweet. We used the trial data  as development data. 

\subsubsection{Model Details}
For subtask A, we trained a classifier by fine-tuning a pre-trained BERT Transformer \cite{devlin2018bert} with a linear layer for text sequence classification on top. 

The input sentences\footnote{In BERT, a ``sentence" can be a text sequence of arbitrary length. In our case, a ``sentence" refers to a tweet even if it may span multiple linguistic sentences.} were first tokenized with the BERT basic tokenizer
to perform punctuation splitting, lower casing and invalid characters removal. Then this was followed by WordPiece tokenization \citep{wu2016google} to split words into sub-word units, in accordance with the original BERT approach \cite{devlin2018bert}. The maximum sequence length was defined as 80, with shorter sequences padded and longer sequences truncated to this length. The order of the input sequence was represented by the learned positional embeddings. The input representation for each tweet is the sum of these token, segment, and position embeddings. As only one sentence serves as input, only the sentence A embeddings are used as the segment embeddings \cite{devlin2018bert}.

We selected the BERT\textsubscript{base-uncased} as the underlying BERT model. The BERT\textsubscript{base} consists of 12 Transformer blocks, 12 self-attention heads, and 768 hidden dimension with a total parameters of 110M. It was trained on the BookCorpus (800M words) and the English Wikipedia (2,500M words). Though the BERT\textsubscript{large} model was reported to outperform the BERT\textsubscript{base} in a variety of tasks, training and fine tuning BERT\textsubscript{large} was too computationally intensive given the time limit. Thus we used BERT\textsubscript{base} for accelerated training.
The BERT\textsubscript{base} model includes a special classification embedding \texttt{[CLS]} at the beginning of every sentence, and this token in the final layer was extracted as the aggregate sequence representation for the current classification task. Then a linear layer of 768 dimensions was added on top of BERT\textsubscript{base}, using the \texttt{[CLS]} embeddings of the whole input sequence to predict a binary label. Binary cross-entropy was used as the loss function to fine-tune the classifier.

\subsubsection{Implementation}
The neural network was implemented in PyTorch \cite{paszke2017automatic}, and we used the tokenizer, pretrained WordPiece, and positional embeddings and pre-trained BERT from the library \texttt{pytorch-pretrained-bert}\footnote{https://github.com/huggingface/pytorch-pretrained-BERT}. 
Following the recommendation for fine-tuning in the original BERT approach \cite{devlin2018bert}, we trained our classifier with a batch size of 32 for 2 epochs. The dropout probability was set to 0.1 for all layers. Adam optimizer was used with a learning rate of 2e-5. The training was carried out on an Nvidia 1070Ti GPU; only taking about 6 minutes in total.

\subsection{Subtask B: Categorizing Offense Types}
For subtasks B and C, we adopted an SVM classifier. For these two tasks, the BERT classifier performed close to the baseline on the trial data. This could be caused by the limited amount of the training data for these two tasks or inappropriate selection of hyperparameters. Thus, we built a linear SVM classifier to identify the offense type and target. 

Subtask B requires the distinction between targeted and untargeted offense. We used an SVM classifier with selected character $n$ gram features for subtask B. For the trial data of subtask B, the classifier achieved a macro F$_1$ score of 0.5333 and accuracy of 0.5714; both of them considerably higher than the baseline. But since the labels of two classes were accidentally flipped in our submission, our results were not competitive. We also reconstructed test F$_1$ from the flipped confusion matrix. If the labels were not flipped, the test F$_1$ should be 0.5946.

\subsection{Subtask C: Identifying the Target of Abuse}

Subtask C requires the classifier to identify three types of offense target, 'Individual' ('IND'), 'Group'('GRP') and 'Other'('OTH'). 
The training set is rather imbalanced: The minority class OTH constitutes around 10 percent of all the instances, and only occurs once in the trial data.
We originally were planning to use the same approach as for subtasks A. However, experiments on the trial data showed a weak performance.
For this reason, we decided to use  a linear SVM classifier to identify the offense target with three sub-classes since previous studies indicate that SVM classifiers perform well on classification tasks and at par with deep neural networks when features are well selected~\cite{founta2018large,kumar2018benchmarking}. For this classifier, we used the Scikit-learn \cite{pedregosa2011scikit} implementation, and we used only the training data provided by the shared task.

\subsubsection{Model Details}
Given that character-level $n$-gram could reduce the effect of spelling errors and variations in tweets~\cite{schmidt2017survey}, we used a bag of character $n$-grams (with $n$ ranging from 2 to 7 characters) as features in order to characterize the users' language features as robustly as possible. Since in subtask C, we need to identify different types of offense target, we assume that named entity information will be effective for identifying target types. Named entities information was extracted by spaCy, which is based on the entity types from OntoNotes 5 corpus\footnote{\url{https://spacy.io/api/annotation\#named-entities}}. Given that this task aims to identify three types of targets, namely individual, group and other, we  used the named entity information by classifying all the entity types into three major types and counting the number of each type separately. The first type only includes PERSON entities, the second type consists of entity types related to a group sense, for example ORG, NORG, and GPE, and the last type includes all occurrences of the other entity types.

\citet{nobata2016abusive} found that linguistic features such as tweet length, average word length, number of punctuation, number of discourse connectives can be useful for detecting abusive language. In this study, we adopt 9 features from their work. Besides the $n$-gram features, named entity, and linguistic features, we also adopted emoji and emoticons as additional features, which have been shown to be useful in sentiment analysis tasks~\cite{kouloumpis2011twitter,shiha2017effects}. Emoticons are extracted using the s regular expressions by C.\ Potts\footnote{http://sentiment.christopherpotts.net/tokenizing.html \#emoticons}.
We also added three emoji sentiment features, which consist of the positive, negative, and overall sentiment scores based on the Emoji Sentiment Ranking~\cite{novak2015sentiment}.  

We performed feature selection for the $n$-gram features using a filtering approach with information gain, which has proven to be effective in social media sentiment classification \cite{kubler2018use}.

Our final submission is a linear SVM classifier (C=0.1, squared-hinge loss function) with 1000 selected character $n$-grams of length 2-7. 
Adding linguistic and emoji features resulted in small gains on the trial data and was that not considered useful for the official version.

\section{Results}
\label{sec:results}

\begin{table}[t]
\center
\begin{tabular}{l|rr}
\bf System & \bf F$_1$ macro & \bf Accuracy \\ 
\hline
All OFF baseline & 0.2182 & 0.2790 \\
All NOT baseline & 0.4189 & 0.7209 \\
\hline
BERT\textsubscript{base-uncased} & \textbf{0.8136} & 0.8570 \\
\hline
\end{tabular}
\caption{The official UM-IU@LING result for subtask A, in comparison to the baselines.}
\label{tab:results-A-open}
\end{table}

\begin{table}[t]
\center
\begin{tabular}{l|rr}
\bf System & \bf F$_1$ macro & \bf Accuracy \\ 
\hline
All OFF baseline & 0.1934 & 0.2399 \\
All NOT baseline & 0.4319 & 0.7601 \\
\hline
SVM\textsubscript{character-ngram} & 0.8267&0.8782 \\
BERT\textsubscript{base-uncased} & \textbf{0.8388} & 0.8722 \\
BERT\textsubscript{base-cased} & 0.8094 & 0.8500 \\
BERT\textsubscript{base-multiling-unc.} & 0.4300 & 0.7625 \\
BERT\textsubscript{base-multiling-cased} & 0.8179 & 0.8718 \\
\hline
\end{tabular}
\caption{Results on the trial data for subtask A.}
\label{tab:results-Atrial}
\end{table}

\begin{figure}[t]
\centering
\includegraphics[width=0.45\textwidth]{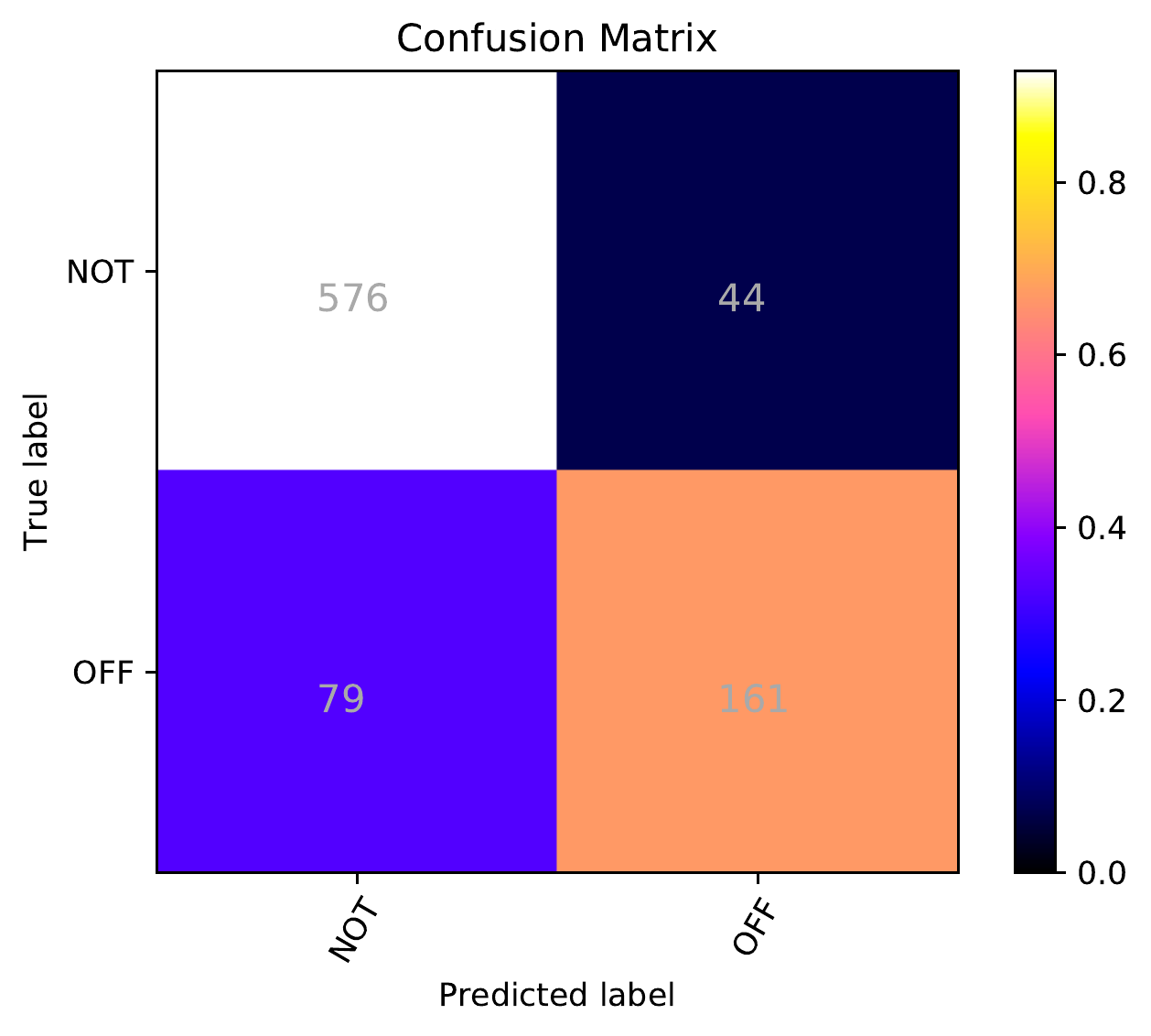}
\caption{The UM-IU@LING confusion matrix for subtask A.}
\label{fig:1}
\end{figure}

\subsection{Subtask A}
Our best result for subtask A along with the official baselines are summarized in Table~\ref{tab:results-A-open}. The BERT classifier achieved a macro F$_1$ score of 0.8136, clearly exceeding the baseline of 0.4189 and ranking the system 3rd out of 103 submissions. This demonstrates that our model can effectively identify whether a given tweet contains offensive content or not. The confusion matrix in Figure~\ref{fig:1} further illustrates the error pattern of our classifier, which more often  misclassified offensive tweets as being not offensive. One explanation of the results may be the classifier's preference for the majority class. But it is possible that our classifier may not capture some of the subtle nuances in meaning and contexts. 
However, the results also show that the macro F$_1$ score is only about 4.5 percent points lower than the accuracy (0.8136 vs.\ 0.8570). This is a clear indication that the classifier is successful in modeling the minority class of offensive tweets.

\begin{table*}[t]
\center
\begin{tabular}{ p{0.7cm} | p{11cm} | p{1cm}| p{1.5cm} } 
 ID & Tweet & Label & Prediction \\ 
    \hline
 50 & okay but it actually sucks so much that the first year I COULD go to every Reeperbahn Festival day, I'm in Strasbourg and can only attend the last day & NOT & OFF \\  \hline
 263 & My mom just called me and said she is joining the NFL boycott. How many of yall are with us? F that league \#NFLBoycott & OFF & NOT \\ \hline
 126 & @User @User @User They don't. The GOP will keep supporting racketeer, illegitimate Trump. They never will stop the corruption of tRump. They are in it for the money. They want to destroy American democracy. & NOT & OFF \\ 
\hline
\end{tabular}
\caption{Misclassified examples for subtask A from the trial data. Usernames are anonymized.}
\label{tab:examples-A}
\end{table*}

\subsubsection{Ablation Analysis}
We performed an ablation analysis on our BERT classifier using the training and the trial data. First, we retrained the classifier by varying the learning rate. The macro F$_1$ dropped to the baseline of 0.4318 with a learning rate of either 2e-8 or 2e-3, which indicates that the system is sensitive to change in learning rates. 

The selection of sequence length only has a minimal influence on the final performance, with a tendency for longer sequence length to improves prediction accuracy: Setting the input sequence length to 60 reduces the macro F$_1$ minimally to 0.8212, and decreasing the  input length to 40 decreases the macro F$_1$ to 0.8126. 

There are several versions of pre-trained BERT\textsubscript{base}\footnote{https://github.com/google-research/bert}. We compared the performance of these different versions of BERT\textsubscript{base} and the results are summarized in Table~\ref{tab:results-Atrial}. Generally, these variants of BERT\textsubscript{base} tend to give similar performance but BERT\textsubscript{base-uncased} achieved the best performance on the trial data. It is unclear why BERT\textsubscript{base-multilingual-uncased} did not learn to perform the task beyond the baseline. Additional hyperparameter tuning might be necessary in this case. Overall, these results demonstrate that though BERT can give superior performance in detecting hate speech, it is somewhat sensitive to the change of hyperparameters. We also find that the SVM classifier achieved a higher accuracy on the trial data, but there is a significant drop in macro F$_1$ when compared with the BERT model. This shows that the BERT model performs better on the minority class. 

\subsubsection{Error Analysis}
We show examples of misclassified tweets in Table~\ref{tab:examples-A}. In example 263, the BERT classifier failed to identify the offensive word ``F".
It is common for people to use euphemisms to tone down  swear words in certain situations. The classifier could miss these word variants, especially when the word variant is the only offensive word in the given tweet. For tweet 50, the word ``sucks'' is the only word that is often used offensively. However, the given tweet is not offensive because the author only describes their mood instead of insulting someone else. These misclassifications seem to indicate that the classifier reacts to trigger words with negative connotations but may not be capable of interpreting the words with respect to the larger context. 

When examining the prediction errors, we consistently noticed that the BERT classifier is highly effective in identifying tweets with words that are negative or offensive in most linguistic contexts. The real challenge is that not all tweets containing negative or potentially insulting words are offensive; there are subtle differences between a negative opinion and an insult towards someone.  However, the model cannot distinguish these subtle differences in meaning in the proper cultural or socio-political contexts. Additionally, it is not robust enough to  detect swear word variants or atypical spellings common in social media.

\subsection{Subtask C}

\begin{table}[t]
\center
\begin{tabular}{l|rr}
\bf System & \bf F$_1$ macro & \bf Accuracy \\ 
\hline
All GRP baseline & 0.1787 & 0.3662 \\
All IND baseline & 0.2130 & 0.4695 \\
All OTH baseline & 0.0941 & 0.1643 \\
\hline
SVM classifier & \textbf{0.5243} & 0.6854 \\
\hline
\end{tabular}
\caption{The official UM-IU@LING results (SVM) for subtask C.}
\label{tab:results-C-open}
\end{table}

\begin{table}[t]
\center
\begin{tabular}{l|rr}
\bf System & \bf F1$_1$ macro & \bf Acc. \\ 
\hline
All IND baseline &0.3041  & 0.8387 \\
All GRP baseline & 0.0762 & 0.1290 \\
All OTH baseline &0.0208 &  0.0323 \\
\hline
SVM\textsubscript{character-ngram} & 0.3915&0.8065 \\
SVM\textsubscript{word-ngram} &0.3554 &0.6774  \\
SVM\textsubscript{char+ling+emoji} & \textbf{0.3971} & 0.8065  \\
SVM\textsubscript{char+ling+emoji+entity} &0.3901 &0.7742  \\
\hline
\end{tabular}
\caption{Results on the trial data for subtask C.}
\label{tab:results-Ctrial}
\end{table}

\begin{table*}[t]
\center
\begin{tabular}{ p{0.7cm} | p{11cm} | p{1cm}| p{1.5cm}   } 
 ID & Tweet & Label & Prediction \\ 
    \hline
17 & @User Obama fed the country shit sandwiches for 8 years. Maybe Jim just has his addled mind confused about dates and who fed who what.. & GRP& IND  \\ \hline
22 & @User The Catholic Church is really screwed up. Nothing new here. & GRP & OTH \\ \hline
31 & @User Yeah thanks to your Nobel Emmy award winning idiot chief flip flopping on everything from Iran to gun control. & IND & GRP \\
\hline
\end{tabular}
\caption{Misclassified examples from the trial data for subtask C.}
\label{tab:examples-C}
\end{table*}

\begin{figure}[t]
\centering
\includegraphics[width=0.45\textwidth]{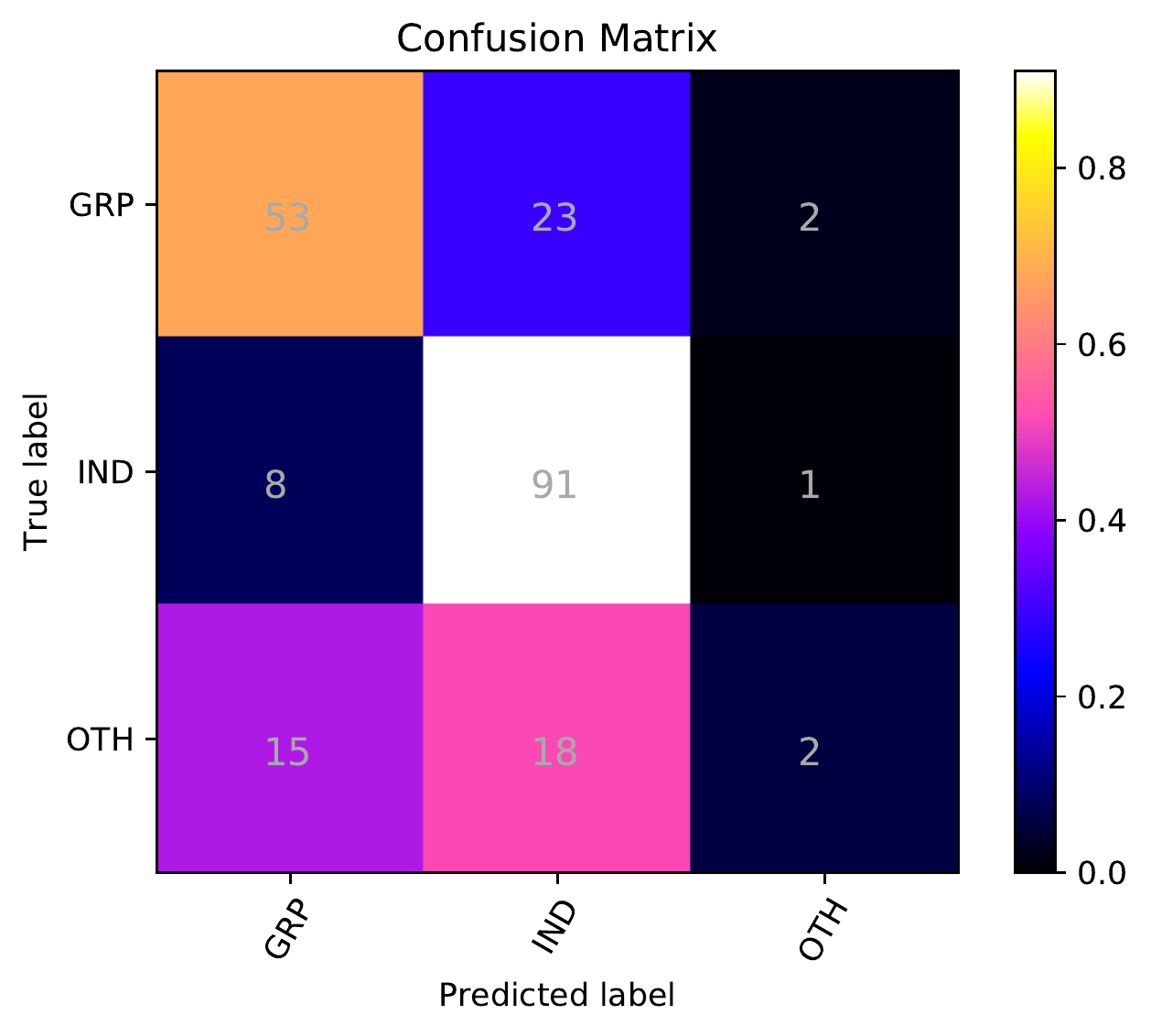}
\caption{Confusion matrix for the SVM classifier for subtask C.}
\label{fig:2}
\end{figure}

Table~\ref{tab:results-C-open} shows our best result for subtask C in comparison to the official baselines. The macro F$_1$ score of the SVM classifier is 0.5243, which is considerably higher than the baseline and ranks the system 27th out of 65 submissions. The confusion matrix in Figure~\ref{fig:2} indicates 
that our classifier performed well on identifying the IND class, was effective for the GRP class, but often failed to distinguish the OTH class from the other two classes. This clearly shows that the sparsity of training data for the minority class OTH affects the performance of our classifier negatively.

The performance of the classifier with different features is shown in Table~\ref{tab:results-Ctrial}. Since there are only 31 instances in the trial set and it is rather imbalanced, we can see that the highest accuracy is reached by classifying all examples as IND, i.e., the all IND baseline.  Even though none of the classifiers outperformed the baseline in terms of accuracy, all the classifiers achieved significantly higher macro F$_1$ scores, which shows that they are better at identifying the other two classes. After adding linguistic and emoji features, the character $n$-gram model showed a slight improvement in macro F$_1$ score and achieved the highest accuracy along with the simple character $n$-gram model. But both macro F$_1$ and accuracy dropped when entity information was added.  

Table~\ref{tab:examples-C} presents examples of misclassified tweets in the trial set. In example 17, two persons are mentioned, ``Obama'' and ``Jim', and both of them are insulted, however not as a group but individually. The classifier labeled this example as IND. In example 22, the classifier is misguided by the word 'Church' and wrongly classifies it as OTH. Example 31 is similar to example 17. Here, there are two potential targets, 'Nobel Emmy award winning idiot' and 'Iran' that could trigger the group sense, which significantly affects the classifier's judgment.

The errors analysis indicates that the classifier has the ability to distinguish individual and group targets, but it fails to capture the relation between different entities and sometimes misidentifies the target category of offensive language.  



\section{Conclusion}
In this study, we report our systems for OffensEval subtasks A and C. In subtask A, we trained a neural network based classifier by fine-tuning the pre-trained BERT\textsubscript{base} model to detect offensive tweets. In subtask C, we used a linear SVM with character $n$-gram features to identify the target of hate speech. 

The evaluation results indicate that our system is capable of detecting offensive language robustly, and it has a good chance of identifying the target. However, there is room for improvement. In the future, in order to capture subtle meaning and overcome the data sparsity, we plan to take syntactic and semantic features into consideration and investigate the combination of selected surface features and pre-trained word embeddings.

\bibliography{semeval}
\bibliographystyle{acl_natbib}

\end{document}